# Biological Robots: Perspectives on an Emerging Interdisciplinary Field

Authors: Blackiston, D.[1,2,4,+], Kriegman, S.[1,2,4,+], Bongard J.[3,4,+], and Levin, M.[1,2,4,+,*]

**Affiliations:**
1 Allen Discovery Center at Tufts University, Medford, MA 02155, USA.
2 Wyss Institute for Biologically Inspired Engineering, Harvard University, Boston, MA 02115, USA.
3 Department of Computer Science, University of Vermont, Burlington, VT 05405, USA.
4 Institute for Computationally Designed Organisms

+ all authors contributed equally to this work.

*Corresponding author: michael.levin@tufts.edu

Email: douglas.blackiston@tufts.edu ; sam.kriegman@northwestern.edu; jbongard@uvm.edu

**Abstract:**

Advances in science and engineering often reveal the limitations of classical approaches initially used to understand, predict, and control phenomena. With progress, conceptual categories must often be re-evaluated to better track recently discovered invariants across disciplines. It is essential to refine frameworks and resolve conflicting boundaries between disciplines such that they better facilitate, not restrict, experimental approaches and capabilities. In this essay, we discuss issues at the intersection of developmental biology, computer science, and robotics. In the context of biological robots, we explore changes across concepts and previously distinct fields that are driven by recent advances in materials, information, and life sciences. Herein, each author provides their own perspective on the subject, framed by their own disciplinary training. We argue that as with computation, certain aspects of developmental biology and robotics are not tied to specific materials; rather, the consilience of these fields can help to shed light on issues of multi-scale control, self-assembly, and relationships between form and function. We hope new fields can emerge as boundaries arising from technological limitations are overcome, furthering practical applications from regenerative medicine to useful synthetic living machines

## Introduction

Multidisciplinary research programs have the potential to generate important advances both within and between established fields; the integration of existing ideas and techniques among researchers with diverse backgrounds often leads to new approaches and new questions. This can be seen in biorobotics, where living materials are being used to build new kinds of robotic devices. Progress in this nascent field will depend on the development of a shared lexicon, as many terms central to biology and robotics lack operational definitions and can have multiple, sometimes incompatible, interpretations when referenced by different fields. For example, what is a robot? Machine? Organoid? Organism? How these terms are used is contingent upon many factors, including disciplinary training, journal readership, reviewers, classical terminology, and the target audience. Debates surrounding biorobotics and its nomenclature can be observed in manuscripts, at conferences, in the popular press, and increasingly on social media. Here, we provide our individual perspectives on the topic and respond to many of the questions raised across disciplines with respect to our recent work on biological robots. The individual viewpoints of our interdisciplinary team are sometimes non-overlapping, yet are integrated toward the overarching goal of improving frameworks for future research, and identifying areas in which dichotomous thinking creates artificial boundaries in understanding.

## Dovetailing developmental biology and robotics (Blackiston commentary)

Biorobotics and materials synthetic biology are relatively young disciplines, with both witnessing a surge in progress across the preceding decade.[1-5] Approaches, model systems, and goals are myriad, combining elements of bioengineering, stem cell biology, molecular biology, computer science, engineering, neuroscience, and robotics. Our work lies at the intersection of these fascinating disciplines, building self-motile biological robots from the ground up from both user and A.I. inspired anatomies, engineered for a specific purpose.[6-8] While these results have engendered broad support from the scientific community and general public, several pointed critiques have been raised by members of the developmental biology community,[9] which are important when framing the work both within the discipline and to those outside the field. As several of these points are valid, the current author offers here their view on the respective topics, which does not necessarily reflect the view of the other co-authors.

The most common point of discussion is that our system is not a robot nor engineered, it is what developmental biologists traditionally define as an animal cap;[10-16] a region of developing frog embryo which gives rise to epidermis and neural tissue in the native system. This position has merit, yet indicates a narrow view of the work and fails to acknowledge the diverse biological robots produced by our design pipeline. Our initial research was comparable to several biohybrid robots, which use a combination of synthetic scaffolds and cardiac muscle-based actuation to generate forward locomotion.[17-21] Using these same muscle-based actuators, our designs replaced the synthetic scaffolds with living materials, creating a combination of modified animal cap and muscle derived tissues.[7] In follow up work, we explored the use of motile cilia-based actuation, small hairlike structures naturally present on the surface of tadpole epidermis.[6,8] The simplest of these designs are indeed spherical animal caps, a point acknowledged in the manuscript, which were used to quantify metrics of velocity and lifespan of ciliated tissues and further advance our computational models. However, the author's view is that this approach represents an advance in materials synthetic biology rather than developmental biology. We leverage tissues and cells as materials to create a sliding scale of design, from fully muscle actuated to fully cilia actuated, as well as combinations of the two, and from non-shaped spheroids to carefully designed anatomies with multiple tissue types placed at specific locations (Fig. 1).

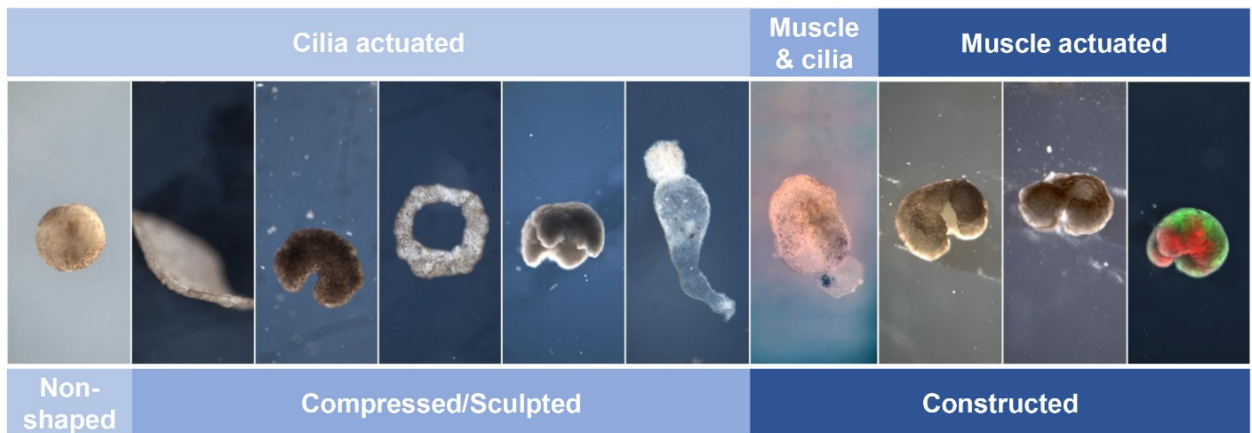

FIG. 1. Sample of designs produced in the research program. Actuation can be achieved through motile cilia generated flow, contractile muscle tissue, or a combination of the two. Morphology can be generated via compression, sculpting, or by layering specific tissue types during the construction process (far right panel, red indicates muscle, green indicates epidermis).

Related to this first point is the idea that the locomotive behavior of animal cap derived tissue is a known phenomenon, a position with which I am in agreement. The fact that ciliated tissues generate sufficient flow to create motion has been reported in the literature for decades,[22-24] and the author would not claim to have discovered something new or unexpected in the system. To the contrary, our results indicate that the cilia and muscle actuators behave like wind-up toys in a predictable and stereotyped fashion, a feature which we leverage with modeling to simulate how individual, and groups, of biological robots will behave *in vitro*. It is precisely through manipulation of these known and stereotyped behaviors that we were able to generate kinematic self-replication,[8] a fragile process requiring careful titration of several input variables from the investigator. Thus, our innovation comes not from novel developmental insight, but rather using these known systems as living materials, enhancing, or changing movement and behavior through anatomical design by placing actuators at specific locations, or by manually layering tissues into novel configurations. A future goal remains better control of the design process to generate motion and behaviors which would not be predicted in traditional animal caps.

Similarly, there has been controversy when using the name xenobots to describe our biological robots, as it represents a rebranding of the established animal cap nomenclature. This author's view is that the term xenobot is a non-technical descriptor and one that was not used in the first manuscript, nor the third following community feedback. However, an umbrella term remains necessary to capture the full design space of the research program, which includes cilia and muscle-based actuation, various tissue geometries, sizes, and in the future, cells derived from diverse taxa. Internally, team members use differing vocabulary to describe the system, including reconfigurable organisms, computer designed organisms, biobots, xenobots, living robots, and biological robots, with the latter being the present author's preferred term as it captures the full range of design space. Additionally, this term should only be applied to the system where motility is used or designed for a pre-specified purpose, and an unmodified animal cap, or the differentiated mucociliary organoid,[23,25-28] should be labeled as such where applicable. Finally, while reconfigurable organism is the specific nomenclature used in the manuscripts (the current author does not consider the designs organisms), this term can also be problematic as biologists currently have no operational definition of the word organism,[29,30] with some arguing the term is without technical meaning.[31]

There has also been dispute when using the term "robot" to describe our designs. Here I must respectfully disagree with some members of the developmental biology community. Across this multidisciplinary research process, the author has come to appreciate that robots are defined by their design principles and function, rather than by the fabrication materials. Indeed, many robots are constructed from non-intuitive components including cardiac powered biohybrid designs,[19,21,32-34] pinecone and oat seed robots driven by hygromorphic actuators,[35,36] liquid droplets,[37-40] and a light driven *C. elegans* RoboWorm,[41] to name but a few. In this framing, there is no difference between a robot constructed from a synthetic soft-scaffold and muscle tissue and one constructed from a living soft-scaffold and muscle tissue. Both are designed, fabricated, and evaluated for performance, and both function through predictable open-loop control systems. These principles and continued exposure to the robotics field have been rewarding and inspiring for the author, who hopes members of other communities will likewise have an open mind when discussing the discipline.

Beyond the immediate science, an important point of discourse surrounding this research relates to the larger issue of science communication, in its many forms and venues.[42] How, and where, should one promote one's own findings to both the scientific community and general public in an increasingly digital age? Our current research has been fortunate to receive significant visibility from the popular press, in both written and video formats. Engaging with these venues remains up to the investigator and there exists potential for both scientific benefit and risk with either course of action. Sensationalism, especially in headlines, drives traffic and revenue for periodicals, and a resistance to commentary with the popular press removes an important check to accurately reporting one's results. Alternatively, engaging with popular press represents a potential conflict of interest, as the investigator is both producing and selling the product, effectively creating a non-reliable narrator and eroding trust in the discipline at large.[43] While the current author maintains caution when engaging with the press, it is also a reality that these venues will generate value as privatized funding sources become increasingly common.

Finally, peer review, both during publication and through subsequent dissemination of the work remains essential to the research process. Presently, we have witnessed this review moving from the realm of grants, manuscripts, and conferences into the digital world in the form of social media. The benefits here are many; investigators can rapidly reach a large community when sharing ideas, troubleshooting methods, creating a professional network, and providing scientific critique. However, there is also danger when scientific critique is built on an emotional response towards an individual rather than rational analysis of a result.[44] As a community, we should continue to assume the best in our colleagues. Should a particular result or finding appear misplaced, one's goal should always be to improve the science rather than tear down an individual, and mentorship remains essential to members at all levels of academic positions. However, it is also necessary that scientists separate speculation from data, and we must absolutely hold our colleagues to the highest standards if we are to maintain public trust in scientific results. The current author has had to learn scientific communication on the fly, falling short on several occasions, and continues to appreciate the community feedback as he develops in this space.

**From strange feet to strange machines (Kriegman commentary)**

In Jewish folklore and Greek mythology, it's relatively easy to create an intelligent robot: Simply combine mud (Golem), or sculpt away ivory (Galatea), into the right shape. Without divine intervention, it's much harder to bring inanimate objects to life. Transforming containers of fluids and elastomers into a wiggling soft robot requires careful design and precise manufacturing.[45] To get such a robot to do something else besides wiggling in place is exceedingly difficult.

So, why not start with living materials instead? In the "xenobots" project,[6-8] that's exactly what we did. Creating a self-powered, self-driving, self-repairing, self-replicating xenobot was not trivial – doing so required careful planning, design, and construction – but it was, in essence, as simple as combining and sculpting the right material into the right shape. In our case, the right material was *Xenopus* cells and the right shape was one that, in computer simulations, maximized the likelihood of generating an interesting nonrandom behavior, such as forward locomotion.

This raises a tricky question: At what point do living tissues become a robot? Indeed there are many different pieces of a developing animal that, when isolated from the host, can move and sense on their own. Parts of animals are not robots. However, once we artificially combine and shape them to render desired behaviors, they are no longer merely animal parts, they become artifacts: artificial yet fully biological robots.

If removed from early frog embryos, a small piece of ectoderm called the animal cap will mature to form a ball of skin covered in motile cilia, which can propel the ball forward through water. Early tadpoles glide around the same way, at the same speed as this explanted tissue. However, there are several ways to deflect (or "program") the system away from its natural behavior. In our work to date, we have shaped these tissues into quadrupeds, bipeds, pyramids, toruses, semi-toroids, and various other non-spherical body shapes (Fig. 1). In some designs, we removed cilia entirely and relied instead on localized patches of cardiac tissue to produce volumetric actuation. Different body shapes, and different admixtures of tissue types, resulted in different (and thus artificial) behaviors.

The resulting xenobots are *autonomous*: they are able to maintain their structure (and thus function) without human intervention and exhibit a diverse array of behavioral repertoires. They are also *adaptive*: xenobots can repair their structure after significant damage. But are they *intelligent*?

The xenobots reported to date have no known mechanism of sensor-motor coordination. All the cells within a xenobot can sense and act and talk to their neighbors, but xenobots as a whole do not exhibit perceptive (sense-guided) behavior: Their movement is consistent with models of blindly actuating bundles of motors (open-loop control). Their healing is consistent with models of blindly adhering bundles of sticky particles. Sometimes xenobots appear to behave perceptively: they suddenly "decide" to turn around, or they become "interested" in an object and "examine" it repeatedly. However, the very same behaviors are manifested by quasi-stable bundles of motors when tipped into a new orientation.[46] This is the problem of ascribing a capacity to an agent solely on the basis of its behavior. But this does not preclude future xenobots from genetic modifications that give rise to sense-guided behavior, action alternatives, memory, learning, and other increasingly cognitive behaviors.

Formally, we have referred to xenobots as reconfigurable organisms, a nod to reconfigurable modular robots[47,48] in which robot modules (like the cells within a frog embryo) can be attached, detached, and rearranged to form new structures (configurations). As biorobotics researchers begin to reconfigure other organisms, the term "xenobot" might be reserved to describe a particular subset of reconfigurable organisms. If biological robots are one day built out of cells harvested from the African elephant (*Loxodonta*) instead of the African clawed frog (*Xenopus* laevis) then perhaps they will be called "loxobots" instead of "xenobots". (This is not to say the choice of "frog modules" was entirely arbitrary. Developing frog eggs, and xenobots derived from them, survive in freshwater without food; mammalian-derived robots such as loxobots would require special media with growth factors and nutrients and precisely tuned $CO_2$ levels.) There is also the possibility of chimeras (mixtures of species) and biohybrids (mixtures of living and artificial materials). Any one of the creatures in this menagerie is just as likely to invoke the literal interpretation of xenobots: "strange robots".

Whether we call them robots or organisms, and whether or not they become chimeras or cyborgs, designing and optimizing such systems is not only strange but extremely non-intuitive. To make the design problem tractable for human minds, the solution space must be winnowed down to a vanishingly small subset of possible forms and functions. Breaking free from design constraints imposed by human cognitive limits would greatly widen our search for useful technologies and new knowledge, but it will by definition require nonhuman assistance. Computational tools (old[7,49-53] and new[8,54-56]) are poised to help.

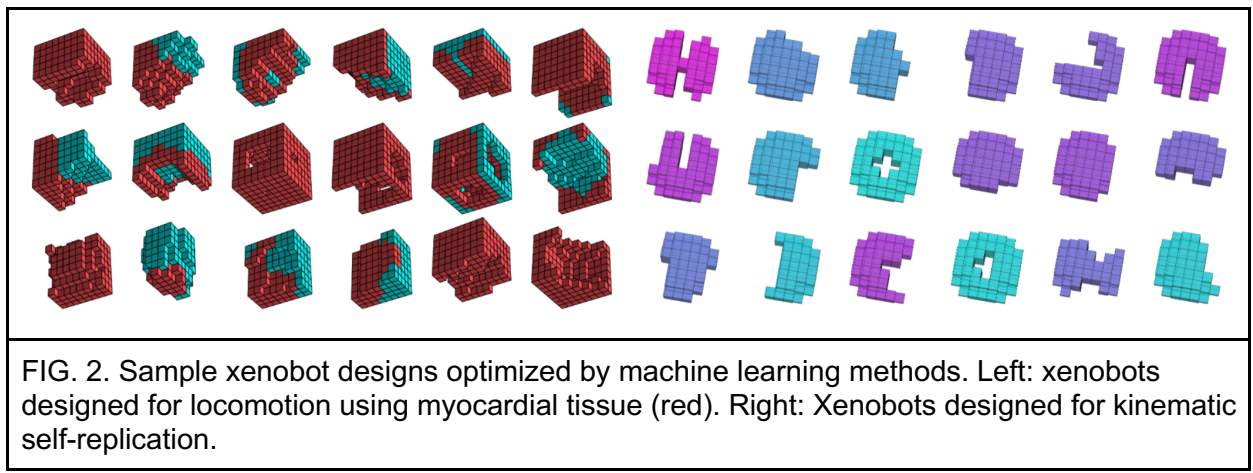

FIG. 2. Sample xenobot designs optimized by machine learning methods. Left: xenobots designed for locomotion using myocardial tissue (red). Right: Xenobots designed for kinematic self-replication.

We used computers to rapidly generate a diversity of buildable xenobot forms that maximized a desired behavior, and maintained that behavior across a range of simulated conditions (Fig. 2). This led us into parts of design space human engineers typically do not wander. There, we not only found creative xenobot designs (fractals, strange asymmetries, and porous structures) but also new biological design principles (e.g. how the right morphology can cohere the noisy actions of its parts).

Exploring this space virtually, allowed us to filter out billions of bad designs before attempting to build them in reality. This greatly improved efficiency in the wetlab and reduced biowaste. But this process can be energy-intensive and its carbon footprint grows with each second of simulation time. This is arguably the most pressing ethical concern of xenobot research. Future work should thus strive to encode xenobot design space such that, instead of meandering through random design variants,[57] gradients can be efficiently followed toward optimal solutions.[58] If successful, this move from trial-and-error to gradient descent could reduce processing time, and thus carbon footprint, by several orders of magnitude.

Biological robots, however efficiently designed, may never become a viable technology. This fact can get lost in the rhetoric surrounding xenobots. When engaging with the popular press and funding agencies (or when trying to explain your research to your grandma) it is helpful to articulate a compelling purpose. We have communicated several visions of the future in which xenobots clean up the environment, monitor industrial processes, deliver medicine, and even fight cancer. It is important to remember that xenobots are not yet useful: they cannot solve any problems at all.

This does not mean that they are useless. Biological robots, like many of the soft robots reported in this journal, challenge conventional views of robotics. They force us to think creatively about how to achieve even the most basic behaviors. The knowledge we gain from our attempts to optimize their design, formalize their control, broaden their functionality, and enhance their intelligence, will eventually become embodied in future useful technologies, whether they are composed of cells, steel, or silicone.

**Expanding robotics by combating dichotomous thinking (Bongard commentary)**

Like every other endeavor, science is not exempt from bearing the marks of the limits of human cognition. Primary among these limits is dichotomous thinking: Making sense of nature, and channeling what we know into embodied machines or disembodied neural networks, requires us to impose distinctions on natural phenomena. However, nature need not, and does not respect human attempts to draw boundaries. What follows are some of the most obvious distinctions the scientific community has attempted to impose on nature, and how technological and biological advances are increasingly corroding them.

In genetics, a clear separation between genotype and phenotype has been imposed ever since genes were first hypothesized and then discovered. However, advances in epigenetics and synthetic biology are increasingly revealing that genes, environment and phenotype are often more usefully thought of as a coupled dynamical system. Xenobots for instance support the idea that a developing organism represents a trajectory through an attractor space. The "default phenotype" usually arrived at in response to environmental signals experienced by the organism in its natural environment is but one attractor in this space. Environmental change, ectopic perturbation, or even AI-designed tissue rearrangements can push a collection of cells into stable adult forms completely different from the one usually observed in nature

In robotics and AI, Cartesian dualism – the West's most famous exemplar of dichotomous thinking – has biased the kinds of scientific questions we ask about intelligence, and the kinds of engineering approaches we take in attempting to create artificial intelligence. Indeed, the very bicameral nature of the robotics and AI communities illustrates how Cartesian dualism warps the field(s). Evidence is now mounting that Cartesian dualism has driven AI into a local optimum:

state-of-the-art non-embodied AI is just as vulnerable to adversarial attacks or out-of-distribution environments as it always was,[59] and calls to incorporate causal reasoning into AI are growing[60] because non-embodied deep networks cannot cause effects and reason about the results.

Our self-replicating xenobots upend a third form of dichotomous thinking that is poisoning computational research: the distinction between tape and machine. This distinction goes all the way back to Turing's original formulation of a theoretical, non-human "computer". However, there is no clear "tape" and "machine" in our self-replicating xenobots: nowhere is there some formal instruction to "find loose cells and push them into copies of yourself". Indeed the xenobots' ability to self-replicate emerges from the collapse of all three forms of dichotomous thinking mentioned above. First, the form and function of the xenobots arise from complex feedback loops between their "genotype" (genetically unmodified frog DNA), phenotype (shape and movement pattern), interoceptive environment (cells and their neighbors within a single xenobot) and exteroceptive environment (the replicative raw material of dissociated cells, and other xenobots) . Second, the "body" of a xenobot is not a binary property during self-replication: there are simply smaller and larger piles of frog cells, and less- or more-motile piles. And, a xenobot's "brain" is the net result of aneural intercellular electric, mechanical, and chemical communication. Third, the geometry of each xenobot dictates how it moves and how, or whether, it contributes to replication: in other words, "the shape is the tape".

Xenobots are thus an ideal guide for leading us, step by step, into the increasingly deep waters of total morphospace. We must leave behind the shallows, where animals and robots with distinct bodies and brains, genotypes and phenotypes, tapes and machines, swim. Such agents do not actually exist. They only exist in our imaginations, because they are the easiest kinds of agents for us to understand. We must leave them and dichotomous thinking behind, and instead learn to swim in the deeps, where real animals reside, and where really intelligent machines will reside. Such animals and machines are ever-shifting patterns of intricately interdependent and fractally-arranged bodies and brains, formal descriptors and physical structures, made that way by the ever shifting currents of natural and artificial selection: it is admixture, all the way down. Such creatures may yield to our understanding, but they will not yield to our attempts to divide them.

**Expanding biology: what biorobots tell us about evolution, morphogenesis, and control (Levin commentary)**

Developmental biology began as a study of the phenomenology of specific embryonic model systems. Modern developmental biology however comprises not only embryogenesis but also regeneration in adult organisms, metamorphosis, and increasingly - digital and synthetic morphogenetic systems.[61-64] What binds these diverse biological examples together into a field is the ability to go beyond specific instances toward questions of multiscale control: how do single cells (which used to be organisms themselves) cooperate toward invariant form and function?[65] How does this process increase complexity while resisting external perturbations and harnessing noise, and how does evolution give rise to self-assembling complex systems with both built-in and learned behaviors?  Exclusive focus on the N=1 example of life provided by Earth's specific phylogenetic tree, with its baggage of frozen accidents, obscures deep principles of life-as-it-can-be.[66] Testing our theories with data outside the dataset of forms that generated them is an essential aspect of any science, crucial not only for possible exobiology but to gain a deeper understanding of evolvability, plasticity, and the relationship between genomically-specified cellular hardware and the physiological software of life. Understanding the rules of life (moving from zoology/botany to true Biology) requires us to recreate and analyze novel living constructs never-before existing on Earth.

Thus, the study of bioengineered forms and their multiscale control policies are an important part of a maturing developmental biology. Xenobots are a good example in which to examine the converse blurring of lines between classical robots and organisms. They challenge us to ask key questions about what we mean by "robot" and "machine", and whether those binary categories really facilitate understanding and progress.[67,68] They challenge us to expand traditional notions of a "program" beyond linear code written by a human to probabilistic, parallel, naturally-evolving control policies embodied in biological components that enable the highly flexible, adaptive interoperability of life seen in chimeras and bio-hybrid constructs.[69] Cells are routinely thought of as molecular machines;[70,71] one of the key challenges of the coming decades is to develop frameworks (or borrow them from information science, cybernetics, computer science, and behavior science) with which to understand the principles by which these machines robustly scale to solve morphogenetic and physiological problems. Developmental and regenerative plasticity[72] clearly reveal that multicellular collectives can handle large degrees of novelty; how does evolution result in robust plasticity, and how can we take advantage of these architectures for improved robotics and artificial intelligence? Xenobots reveal new ways to think about control, design, and the multiscale behavior across the spectrum of natural and artificial systems[73-75].

Are Xenobots engineered or natural products of what Xenopus cells do already? Yes, both. Engineering is not just about adding new ingredients, such as DNA plasmids for synthetic biology circuits or nanomaterials (although these will certainly be added in the future). Xenobots reveal a new strategy for the bioengineer's toolbox: programming desired form and function by releasing natural constraints. In Xenobots, nothing was added; instead, cells were liberated from the normal constraints of the rest of the embryo. This led to the discovery that cells' baseline capability is not merely to form a boundary that keeps pathogens out of the organism. Instead, this mundane 2-dimensional lifestyle is forced on them by the instructive interactions of other cell types. When allowed to express their own multicellularity, these epithelial cells do not die, form a monolayer, or wander aimlessly. Instead, their baseline behavior is to form a self-motile proto-organism with individual and collective behaviors such as kinematic self-replication.[8]

These types of form and function were not obvious from any first principles or from their wild-type genotype, and illustrate the need to understand what it is that the evolutionary process taught the Xenopus laevis genome;[76,77] it was not only how to make a frog or a tadpole. There has never been selection to form a functional Xenobot, which raises the important question of what defines the class of constructs that cells can make, beyond the default configuration supported by eons of selection forces. What else do these, and all the other, cells know how to do, in novel circumstances? Of course, their capabilities are explainable, after their discovery, as genetically-specified features (such as cilia and adhesion proteins) working in novel ways through the laws of physics. We encourage readers to make (and register in advance) predictions of other capabilities that cells such as these will exhibit - the degree of their plasticity and the specific forms and behaviors of which they might be capable. The predictions of their capacities and limitations (so-called developmental constraints) will be interesting to test against the rapidly emerging data in this field. The fact that these are no more predictable *ab initio* from genomic information and current models than the primary shapes of animals and plants is an important challenge to the field.

The Xenobots' ability to assemble the next generation from cellular material in their environment sheds an important light on the notion of control, both in the evolutionary and in the engineering context. Indeed, both use the same strategy: reliance on the competency of their components. Engineers make the first generation of Xenobots by relying on the willingness of cells to get together despite novel circumstances and create a motile, coherent agent. These Xenobots create their next generation by exactly the same process: they assemble the cells and impact the size and composition of the collective, but then take advantage of the competency of cells to do the rest - guided self-assembly (i.e., behavior shaping), not micromanagement is the way that we, and the Xenobots themselves, make more Xenobots. Indeed, evolution itself relies on the competency of cells, tissues, and organs to solve problems.[78]

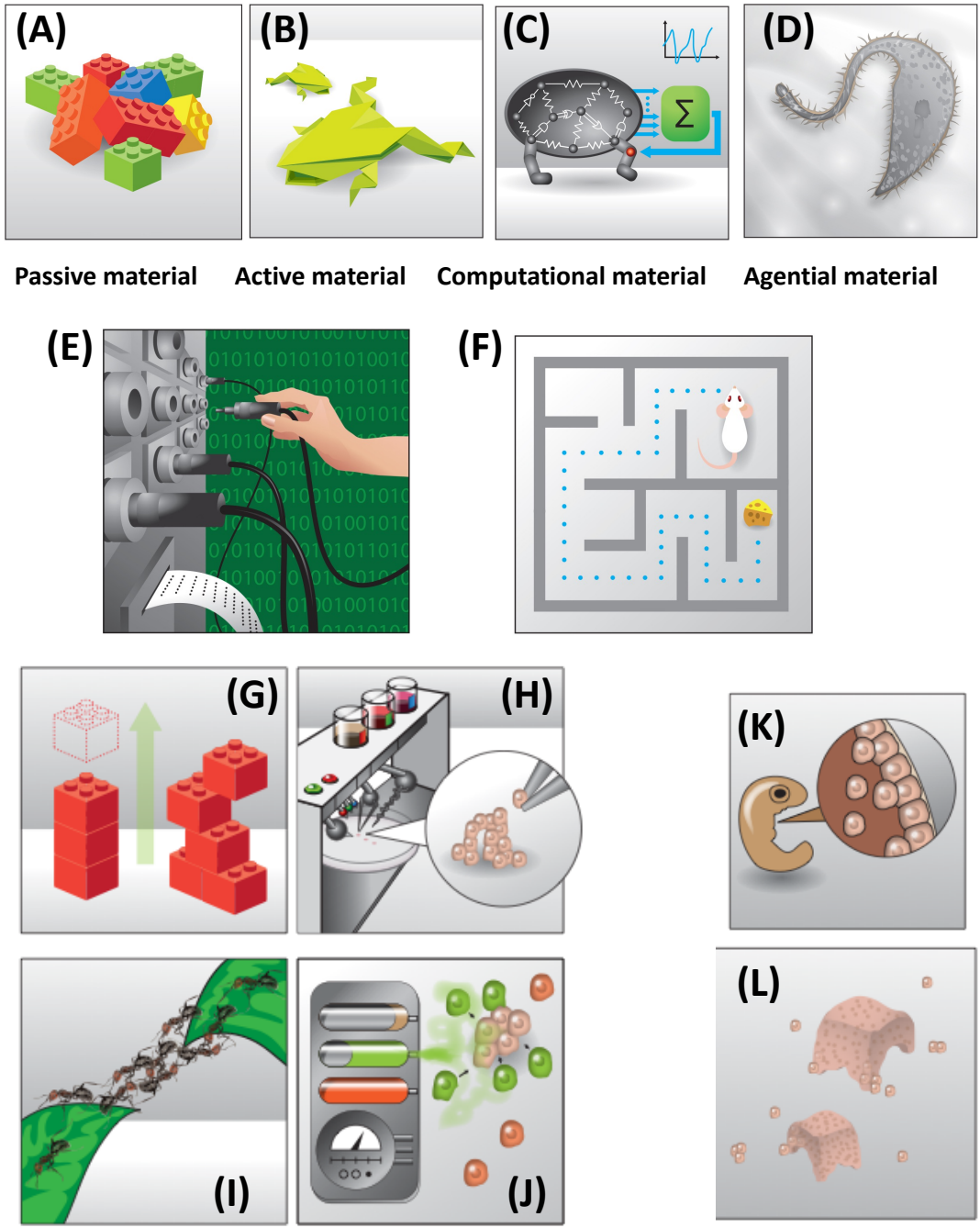

FIG 3: engineering with agential materials. Engineering has traditionally been done with passive materials (A) which can only be dependent upon to keep their shape and other physical properties. These must be carefully managed for each desired functionality, giving rise to a perception of robotics as the manual arrangement of parts toward each goal. However, increasingly engineering has moved toward active matter (B) and computational media (C) as well-recognized in soft robotics[79-81]; now, biorobotics enters a new phase transition where the material is "agential" – it is composed of subunits (D, living cells) which themselves were whole organisms once and thus have many built-in competencies and agendas[82], including problem-solving in physiological, metabolic, and morphological problem spaces[78,83]. This means that robots are now not only constructed by physical (or even genetic) rewiring (E) but more akin to behavior-shaping (F), using signals and environments to achieve desired system-level behavior. In contrast and complement to 3D-printing and similar approaches designed for building with passive matter (G), which also works with cells (H), collective intelligence of living systems at all scales (such as that of an ant swarm, I) can be used to manipulate the collective behavior of cells (J) in anatomical morphospace, by signaling that alters the collective's navigation policy of that space: just as instructive signals from other cells cause frog ectodermal cells to be a 2-dimensional barrier in standard embryos (K), techniques such as subtraction (of other cells and their signals) and stimuli can achieve guided self-assembly toward novel form and function (L).

Making a copy of a biobot does not have to be micromanaged any more than the original biobot needed to be 3-D printed. This does not resemble traditional robotics because the history of engineering has relied on *passive* materials. Progress is revealing a continuum of control strategies appropriate to active, computational, and *agential materials*: building things with parts that have context-sensitive activity, computational capacity, and behavioral agendas that optimize various parameters in local problem spaces. Evolution likewise relies on its components to provide more than reliable passive form (LEGOs that hold their shape but do nothing else) - instead, it largely shapes signals and biophysically-implemented incentives to modulate the behavior of cells, tissues, and organs. Biologists, roboticists, and intellectual property lawyers will have to become comfortable with workflows in which we are literally collaborators with our artifacts, parts of which will actively build the next level of the construct in ways that are not fully captured by a craftsman's action protocol. This has massive advantages, for rational design and for evolution, because it allows both to work in a simpler space of incentives and inputs, not microstates. It also provides extensive challenges, because our science of prediction and control of systems with a multi-scale competencies architecture is still nascent.

The existing variants of Xenobots have not yet been provided with edited genomes, new materials, etc. Counter to established intuitions, the kind of control expected of robotic platforms (machines) does not require genomic editing. Computer science achieved the information technology revolution by moving from programming by hardware rewiring to software control via inputs. Evolution discovered this trick via physiological software very early on,[84] and recent advances in the control of growth and form show how much anatomical change can be implemented by interventions that do not change the underlying hardware (genetics).[85] Induction of appendage regeneration, production of structures belonging to other species despite a wild-type genome, induction or normalization of cancer, etc. can all be induced by biophysical information signals and often invisible to canonical molecular biology tools.[86] A key aspect of the early generations of Xenobots is precisely that they reveal the plasticity of cells' ability to explore morphospace and behavioral space with the same genome. This is an expansion of "epigenetics", toward a better understanding of plasticity and the actual products of evolution.

The practical applications of such biobots are numerous. Major advantages are their biodegradable, soft nature and the built-in competencies of cells for sensing, metabolic/biochemical activity, and actuation in ways that are far beyond today's engineering efforts. Most obvious are the useful living machines which could microsculpt bioengineered tissues in vitro, perform sensing or cleanup in many industrial processes, environmental cleanup, or even in-body biomedical applications with respect to injury, infectious microbes, and cancer.

They also provide an inexpensive, safe model system for education in STEM efforts (such as Frugal Science etc.). Deeper impacts include the use of this platform as a sandbox in which to crack the morphogenetic code and move closer to the endgame of an "anatomical compiler". Work in biobots will help address the gaps in our ability to predict and control complex anatomical shapes[65,87] for applications in developing signaling protocols to induce cells in the body to achieve correct regenerative morphogenesis. An essential aspect of such strategies to address birth defects, traumatic injury, and errors of multicellularity such as cancer and aging, is learning to control the shape toward which cellular collectives cooperate. Biobots are an enabling platform to achieve Feynman's dictum - to truly understand morphogenetic systems by building them ourselves.

Benefits will also accrue to computer science and robotics in inorganic media because the fascinating features of biobots do not derive from any magic of protoplasm. Rather, they derive from their multiscale competency architecture and other principles to be discovered. Current robotics is safe from cancer (defections of components from system-level goals) but is brittle and lacking general intelligence. The opportunity is to borrow from nature, not the specific genes and pathways comprising today's developmental biology textbooks, but the ways in which non-neural systems solve problems in diverse problem spaces by scaling their basal homeostatic properties into larger goals.[88,89] The implications for AI and behavioral science in general concern biobots' degrees of proto-cognitive sophistication. No claims have yet been made for the degree of agency they can exhibit (this characterization is currently under way), but it is clear that humans' intuitions about recognizing intelligent behavior are highly limited by the training set of familiar animals behaving in 3D space.[78,90] The many surprises in the basal cognition literature[91] reveal that we are fundamentally not good at recognizing possible intelligence in navigating physiological, morphological, and other non-obvious spaces; a rich research program exists around efforts to understand Xenobots' unconventional intelligence (at the very least, as a warm-up to exobiological and AI advances). Finally, the long-term implications of this research program address a key component of avoiding existential risk for humanity: better learning to understand and predict the behaviors of systems composed of competent components (emergent properties of collective intelligences).

This technology also has important implications for ethics. These are not limited to familiar biosafety concerns, as these technologies are much easier to contain and have much better safety profiles than existing efforts in synthetic bacteria, viruses, and genetically-modified crops and other organisms. Xenobots for example live only in very specific environments and are highly dependent on humans for their manufacture; they biodegrade rapidly - unlike many of today's synthetic biology efforts that replicate readily in the biosphere. An aspect often forgotten is that we cannot just evaluate potential *risk* as if the status quo was excellent, and our goal was merely not to make things worse. Human and animal suffering world-wide is enormous, and due to a poor understanding of system-level biological control. There is a massive opportunity cost of not pursuing this research, because of its potential to advance regenerative medicine. Any risk calculations of such technologies must balance against the moral duty to pursue scientific programs that can alleviate unmet biomedical needs and health care disparities. We ethically owe it to victims of birth defects, injury, cancer, and degenerative disease etc. to better understand how to influence cells to achieve complex structural and functional states.[92]

**Conclusion**

Robotics includes the engineering of useful, partly- or fully autonomous artifacts that remind us in some way of organisms. Increasingly, this research is breaking with the classical conception of a robot as necessarily made of passive inorganic components and made or designed piece-by-piece by humans at every step. The essential nature of robotics is not limited to a specific kind of material, origin story, or type of control. However, this also necessitates a re-evaluation of terminology as multidisciplinary teams bring field specific techniques to bear on new questions. As we lay out, this terminology remains in a state of flux, as audiences continue to view these questions through their own unique lens.

We suggest that many new research programs challenge the traditional lines drawn between a robot and an organism. Living, "active", "smart", or "agential" materials (depending on the author) have much to offer robots and engineering disciplines. Simultaneously, materials science research can provide insights into basic developmental biology and bioengineering as cells and tissues assemble during the design process. From the opposite end, simulation and AI-driven design have the potential to enhance traditional biology programs, as sim-to-real approaches continue to accelerate *in vivo* and *in vitro* discovery. Within the group, the authors vary in specific terminology and system label usage (e.g., robot vs. organism vs. xenobot), whether they characterize designs as a group of skin cells or something new, and when speculating about the degree of sensory motor communication possible in the system. These issues remain areas of active investigation. Finally, we understand where disagreements arise among fields and hope the views presented here help frame the research program to the larger scientific community.

Many fields of science and technology, as they mature, go beyond the specific matter of their early applications. Engineering is no longer limited to the study of engines. Just as the deep principles of physics and thermodynamics apply across scales, from galaxies to the soft matter of cellular processes, biology is moving from zoology to unraveling the fundamental laws of life across natural, synthetic, and exo-biological systems. Computer science and robotics can lead the way, with their early emphasis on functionality and realizability, not material implementation. Thus, importing the ideas of programmability and substrate independence will open further opportunities in the life sciences and bioengineering, to go beyond the contingent natural products of evolution to truly understand "life as it can be".[93] Together, the emerging consilience of techniques and concepts at the intersection of materials science, computing, and bioengineering is an exciting new field with the potential the create of technological machines that embody biological principles, and incorporate biological components, in entirely new ways.

**Acknowledgements.** This work was sponsored by the Defense Advanced Research Projects Agency (DARPA) under Cooperative Agreement Number HR0011-18-2-0022, the Lifelong Learning Machines program from DARPA/MTO. The content of the information does not necessarily reflect the position or the policy of the government, and no official endorsement should be inferred. Approved for public release; distribution is unlimited. This research was also supported by the Allen Discovery Center program through The Paul G. Allen Frontiers Group (12171), and M.L. gratefully acknowledges support from the National Science Foundation's Emergent Behaviors of Integrated Cellular Systems Grant (Subaward CBET-0939511). This research was also supported by the National Science Foundation's Emerging Frontiers in

**References**

1. Tang, T.-C. et al. Materials design by synthetic biology. *Nature Reviews Materials* **6**, 332-350 (2021).
2. Ricotti, L. et al. Biohybrid actuators for robotics: A review of devices actuated by living cells. *Sci Robot* **2**, doi:10.1126/scirobotics.aaq0495 (2017).
3. Sun, L. et al. Biohybrid robotics with living cell actuation. *Chemical Society Reviews* **49**, 4043-4069 (2020).
4. Mestre, R., Patino, T. & Sanchez, S. Biohybrid robotics: From the nanoscale to the macroscale. *Wiley interdisciplinary reviews. Nanomedicine and nanobiotechnology* **13**, e1703, doi:10.1002/wnan.1703 (2021).
5. Heinrich, M. K. et al. Constructing living buildings: a review of relevant technologies for a novel application of biohybrid robotics. *Journal of the Royal Society, Interface* **16**, 20190238, doi:10.1098/rsif.2019.0238 (2019).
6. Blackiston, D. et al. A cellular platform for the development of synthetic living machines. *Science Robotics* **6**, eabf1571 (2021).
7. Kriegman, S., Blackiston, D., Levin, M. & Bongard, J. A scalable pipeline for designing reconfigurable organisms. *Proceedings of the National Academy of Sciences* **117**, 1853-1859 (2020).
8. Kriegman, S., Blackiston, D., Levin, M. & Bongard, J. Kinematic self-replication in reconfigurable organisms. *Proceedings of the National Academy of Sciences* **118** (2021).
9. Ratcliff, W. C. The biological robots are coming! But note they have been here for ~3.5 billion years. *GEN Biotechnology* **1**, 26-27 (2022).
10. Green, J. The animal cap assay. *Methods in molecular biology* **127**, 1-13, doi:10.1385/1-59259-678-9:1 (1999).
11. Ariizumi, T. et al. Isolation and differentiation of Xenopus animal cap cells. *Current protocols in stem cell biology* **Chapter 1**, Unit 1D 5, doi:10.1002/9780470151808.sc01d05s9 (2009).
12. Gallagher, B. C., Hainski, A. M. & Moody, S. A. Autonomous differentiation of dorsal axial structures from an animal cap cleavage stage blastomere in Xenopus. *Development* **112**, 1103-1114 (1991).
13. Sive, H. L., Grainger, R. M. & Harland, R. M. Animal Cap Isolation from Xenopus laevis. *CSH protocols* **2007**, pdb prot4744, doi:10.1101/pdb.prot4744 (2007).
14. Sasai, Y., Lu, B., Piccolo, S. & De Robertis, E. M. Endoderm induction by the organizer-secreted factors chordin and noggin in Xenopus animal caps. *The EMBO journal* **15**, 4547-4555 (1996).
15. Jones, E. & Woodland, H. The development of animal cap cells in Xenopus: a measure of the start of animal cap competence to form mesoderm. *Development* **101**, 557-563 (1987).
16. Sokol, S. & Melton, D. Pre-existent pattern in Xenopus animal pole cells revealed by induction with activin. *Nature* **351**, 409-411 (1991).
17. Webster, V. A. et al. in *Conference on Biomimetic and Biohybrid Systems.* 365-374 (Springer).
18. Webster, V. A. et al. in *Conference on Biomimetic and Biohybrid Systems* 475-486 (Springer, Cham).
19. Sakar, M. S. et al. Formation and optogenetic control of engineered 3D skeletal muscle bioactuators. *Lab on a chip* **12**, 4976-4985 (2012).
20. Lin, Z., Jiang, T. & Shang, J. The emerging technology of biohybrid micro-robots: a review. *Bio-Design and Manufacturing*, 1-26 (2021).

21   Cvetkovic, C. *et al.* Three-dimensionally printed biological machines powered by skeletal muscle. *Proc Natl Acad Sci U S A* **111**, 10125-10130, doi:10.1073/pnas.1401577111 (2014).
22   Angerilli, A., Smialowski, P. & Rupp, R. A. The Xenopus animal cap transcriptome: building a mucociliary epithelium. *Nucleic acids research* **46**, 8772-8787, doi:10.1093/nar/gky771 (2018).
23   Kang, H. J. & Kim, H. Y. Mucociliary Epithelial Organoids from Xenopus Embryonic Cells: Generation, Culture and High-Resolution Live Imaging. *Journal of visualized experiments : JoVE*, doi:10.3791/61604 (2020).
24   Huynh, M. H., Hong, H., Delovitch, S., Desser, S. & Ringuette, M. Association of SPARC (osteonectin, BM-40) with extracellular and intracellular components of the ciliated surface ectoderm of Xenopus embryos. *Cell motility and the cytoskeleton* **47**, 154-162, doi:10.1002/1097-0169(200010)47:2<154::AID-CM6>3.0.CO;2-L (2000).
25   Werner, M. E. & Mitchell, B. J. Using Xenopus skin to study cilia development and function. *Methods in enzymology* **525**, 191-217, doi:10.1016/B978-0-12-397944-5.00010-9 (2013).
26   Walentek, P. & Quigley, I. K. What we can learn from a tadpole about ciliopathies and airway diseases: Using systems biology in Xenopus to study cilia and mucociliary epithelia. *Genesis* **55**, doi:10.1002/dvg.23001 (2017).
27   Kim, H. Y., Jackson, T. R., Stuckenholz, C. & Davidson, L. A. Tissue mechanics drives regeneration of a mucociliated epidermis on the surface of Xenopus embryonic aggregates. *Nat Commun* **11**, 665, doi:10.1038/s41467-020-14385-y (2020).
28   Walentek, P. Manipulating and Analyzing Cell Type Composition of the Xenopus Mucociliary Epidermis. *Methods in molecular biology* **1865**, 251-263, doi:10.1007/978-1-4939-8784-9_18 (2018).
29   Pepper, J. W. & Herron, M. D. Does biology need an organism concept? *Biological Reviews* **83**, 621-627 (2008).
30   Folse III, H. J. & Roughgarden, J. What is an individual organism? A multilevel selection perspective. *The Quarterly review of biology* **85**, 447-472 (2010).
31   Wilson, J. A. Ontological butchery: organism concepts and biological generalizations. *Philosophy of Science* **67**, S301-S311 (2000).
32   Morimoto, Y., Onoe, H. & Takeuchi, S. Biohybrid robot powered by an antagonistic pair of skeletal muscle tissues. *Sci Robot* **3**, doi:10.1126/scirobotics.aat4440 (2018).
33   Nawroth, J. C. *et al.* A tissue-engineered jellyfish with biomimetic propulsion. *Nature biotechnology* **30**, 792-797, doi:10.1038/nbt.2269 (2012).
34   Park, S. J. *et al.* Phototactic guidance of a tissue-engineered soft-robotic ray. *Science* **353**, 158-162, doi:10.1126/science.aaf4292 (2016).
35   Tamaru, J., Yui, T. & Hashida, T. Autonomously moving pine-cone robots: Using pine cones as natural hygromorphic actuators and as components of mechanisms. *Artificial life* **26**, 80-89 (2020).
36   Ochi, K. & Matsumoto, M. Non-electrically driven robot composed of oat seeds with awns. *Artificial Life and Robotics* **26**, 442-449 (2021).
37   Čejková, J., Banno, T., Hanczyc, M. M. & Štěpánek, F. Droplets as liquid robots. *Artificial life* **23**, 528-549 (2017).
38   Fan, X., Sun, M., Sun, L. & Xie, H. Ferrofluid droplets as liquid microrobots with multiple deformabilities. *Advanced Functional Materials* **30**, 2000138 (2020).
39   Li, F. *et al.* Liquid metal droplet robot. *Applied Materials Today* **19**, 100597 (2020).
40   Fan, X., Dong, X., Karacakol, A. C., Xie, H. & Sitti, M. Reconfigurable multifunctional ferrofluid droplet robots. *Proceedings of the National Academy of Sciences* **117**, 27916-27926 (2020).

| | |
|---|---|
| 41 | Dong, X. *et al.* Toward a living soft microrobot through optogenetic locomotion control of Caenorhabditis elegans. *Science Robotics* **6**, eabe3950 (2021). |
| 42 | Bubela, T. *et al.* Science communication reconsidered. *Nature biotechnology* **27**, 514-518 (2009). |
| 43 | Weingart, P. & Guenther, L. Science communication and the issue of trust. *Journal of Science communication* **15**, C01 (2016). |
| 44 | Cheplygina, V., Hermans, F., Albers, C., Bielczyk, N. & Smeets, I. Ten simple rules for getting started on Twitter as a scientist. *PLoS computational biology* **16**, doi:ARTN e100751310.1371/journal.pcbi.1007513 (2020). |
| 45 | Wehner, M. *et al.* An integrated design and fabrication strategy for entirely soft, autonomous robots. *nature* **536**, 451-455 (2016). |
| 46 | Kriegman, S. *et al.* in *2020 3rd IEEE International Conference on Soft Robotics (RoboSoft).* 359-366 (IEEE). |
| 47 | Yim, M. *et al.* Modular self-reconfigurable robot systems [grand challenges of robotics]. *IEEE Robotics & Automation Magazine* **14**, 43-52 (2007). |
| 48 | White, P., Zykov, V., Bongard, J. C. & Lipson, H. in *Robotics: Science and Systems.* 161-168 (Citeseer). |
| 49 | Sims, K. Evolving 3D morphology and behavior by competition. *Artificial life* **1**, 353-372 (1994). |
| 50 | Jakobi, N. Evolutionary robotics and the radical envelope-of-noise hypothesis. *Adaptive behavior* **6**, 325-368 (1997). |
| 51 | Lipson, H. & Pollack, J. B. Automatic design and manufacture of robotic lifeforms. *nature* **406**, 974-978 (2000). |
| 52 | Cheney, N., Bongard, J., SunSpiral, V. & Lipson, H. Scalable co-optimization of morphology and control in embodied machines. *Journal of The Royal Society Interface* **15**, 20170937 (2018). |
| 53 | Bongard, J., Zykov, V. & Lipson, H. Resilient machines through continuous self-modeling. *Science* **314**, 1118-1121 (2006). |
| 54 | Pathak, D., Lu, C., Darrell, T., Isola, P. & Efros, A. A. Learning to control self-assembling morphologies: a study of generalization via modularity. *Advances in Neural Information Processing Systems* **32** (2019). |
| 55 | Ha, D. Reinforcement learning for improving agent design. *Artificial life* **25**, 352-365 (2019). |
| 56 | Ma, P. *et al.* Diffaqua: A differentiable computational design pipeline for soft underwater swimmers with shape interpolation. *ACM Transactions on Graphics (TOG)* **40**, 1-14 (2021). |
| 57 | Bongard, J. C. Evolutionary robotics. *Communications of the ACM* **56**, 74-83 (2013). |
| 58 | Hu, Y. *et al.* in *2019 International conference on robotics and automation (ICRA).* 6265-6271 (IEEE). |
| 59 | Powers, J., Grindle, R., Frati, L. & Bongard, J. A good body is all you need: avoiding catastrophic interference via agent architecture search. *arXiv preprint arXiv:2108.08398* (2021). |
| 60 | Pearl, J. The seven tools of causal inference, with reflections on machine learning. *Communications of the ACM* **62**, 54-60 (2019). |
| 61 | Tomoda, K. & Kime, C. Synthetic embryology: Early mammalian embryo modeling systems from cell cultures. *Dev Growth Differ* **63**, 116-126, doi:10.1111/dgd.12713 (2021). |
| 62 | Rosado-Olivieri, E. A. & Brivanlou, A. H. Synthetic by design: Exploiting tissue self-organization to explore early human embryology. *Dev Biol* **474**, 16-21, doi:10.1016/j.ydbio.2021.01.004 (2021). |

63  Haase, K. & Freedman, B. S. Once upon a dish: engineering multicellular systems. *Development* **147**, doi:10.1242/dev.188573 (2020).
64  Ebrahimkhani, M. R. & Levin, M. Synthetic living machines: A new window on life. *iScience* **24**, 102505, doi:10.1016/j.isci.2021.102505 (2021).
65  Pezzulo, G. & Levin, M. Top-down models in biology: explanation and control of complex living systems above the molecular level. *J R Soc Interface* **13**, doi:10.1098/rsif.2016.0555 (2016).
66  Langton, C. G. *Artificial life : an overview*.  (MIT Press, 1995).
67  Bongard, J. & Levin, M. Living Things Are Not (20th Century) Machines: Updating Mechanism Metaphors in Light of the Modern Science of Machine Behavior. *Frontiers in Ecology and Evolution* **9**, doi:10.3389/fevo.2021.650726 (2021).
68  Nicholson, D. J. Is the cell really a machine? *J Theor Biol* **477**, 108-126, doi:10.1016/j.jtbi.2019.06.002 (2019).
69  Nanos, V. & Levin, M. Multi-scale Chimerism: An experimental window on the algorithms of anatomical control. *Cells Dev* **169**, 203764, doi:10.1016/j.cdev.2021.203764 (2021).
70  Davidson, L. A. Epithelial machines that shape the embryo. *Trends Cell Biol* **22**, 82-87, doi:10.1016/j.tcb.2011.10.005 (2012).
71  Kamm, R. D. & Bashir, R. Creating living cellular machines. *Ann. Biomed. Eng.* **42**, 445-459, doi:10.1007/s10439-013-0902-7 (2014).
72  Clawson, W. P. & Levin, M. Endless Forms Most Beautiful: teleonomy and the bioengineering of chimeric and synthetic organisms. *Biological Journal of the Linnean Society* **in press** (2022).
73  Davies, J. A. & Glykofrydis, F. Engineering pattern formation and morphogenesis. *Biochem Soc Trans* **48**, 1177-1185, doi:10.1042/BST20200013 (2020).
74  Sole, R., Moses, M. & Forrest, S. Liquid brains, solid brains. *Philos Trans R Soc Lond B Biol Sci* **374**, 20190040, doi:10.1098/rstb.2019.0040 (2019).
75  Sole, R. Synthetic transitions: towards a new synthesis. *Philos Trans R Soc Lond B Biol Sci* **371**, 20150438, doi:10.1098/rstb.2015.0438 (2016).
76  Kouvaris, K., Clune, J., Kounios, L., Brede, M. & Watson, R. A. How evolution learns to generalise: Using the principles of learning theory to understand the evolution of developmental organisation. *PLoS computational biology* **13**, e1005358, doi:10.1371/journal.pcbi.1005358 (2017).
77  Watson, R. A. & Szathmary, E. How Can Evolution Learn? *Trends Ecol Evol* **31**, 147-157, doi:10.1016/j.tree.2015.11.009 (2016).
78  Fields, C. & Levin, M. Competency in Navigating Arbitrary Spaces as an Invariant for Analyzing Cognition in Diverse Embodiments. *Entropy (Basel)* **24**, doi:10.3390/e24060819 (2022).
79  Kaspar, C., Ravoo, B. J., van der Wiel, W. G., Wegner, S. V. & Pernice, W. H. P. The rise of intelligent matter. *Nature* **594**, 345-355, doi:10.1038/s41586-021-03453-y (2021).
80  Pishvar, M. & Harne, R. L. Foundations for Soft, Smart Matter by Active Mechanical Metamaterials. *Adv Sci (Weinh)* **7**, 2001384, doi:10.1002/advs.202001384 (2020).
81  Pfeifer, R., Iida, F. & Lungarella, M. Cognition from the bottom up: on biological inspiration, body morphology, and soft materials. *Trends Cogn. Sci.* **18**, 404-413, doi:10.1016/j.tics.2014.04.004 (2014).
82  Ratcliff, W. C., Herron, M., Conlin, P. L. & Libby, E. Nascent life cycles and the emergence of higher-level individuality. *Philos Trans R Soc Lond B Biol Sci* **372**, doi:10.1098/rstb.2016.0420 (2017).
83  Baluška, F. & Levin, M. On Having No Head: Cognition throughout Biological Systems. *Front Psychol* **7**, 902, doi:10.3389/fpsyg.2016.00902 (2016).

84  Fields, C., Bischof, J. & Levin, M. Morphological Coordination: A Common Ancestral Function Unifying Neural and Non-Neural Signaling. *Physiology (Bethesda)* **35**, 16-30, doi:10.1152/physiol.00027.2019 (2020).
85  Sullivan, K. G., Emmons-Bell, M. & Levin, M. Physiological inputs regulate species-specific anatomy during embryogenesis and regeneration. *Commun Integr Biol* **9**, e1192733, doi:10.1080/19420889.2016.1192733 (2016).
86  Levin, M. Bioelectric signaling: Reprogrammable circuits underlying embryogenesis, regeneration, and cancer. *Cell* **184**, 1971-1989, doi:10.1016/j.cell.2021.02.034 (2021).
87  Pezzulo, G. & Levin, M. Re-membering the body: applications of computational neuroscience to the top-down control of regeneration of limbs and other complex organs. *Integr Biol (Camb)* **7**, 1487-1517, doi:10.1039/c5ib00221d (2015).
88  Levin, M. Life, death, and self: Fundamental questions of primitive cognition viewed through the lens of body plasticity and synthetic organisms. *Biochemical and Biophysical Research Communications* **564**, 114-133, doi:https://doi.org/10.1016/j.bbrc.2020.10.077 (2020).
89  Levin, M. The Computational Boundary of a "Self": Developmental Bioelectricity Drives Multicellularity and Scale-Free Cognition. *Front Psychol* **10**, 2688, doi:10.3389/fpsyg.2019.02688 (2019).
90  Levin, M. Technological Approach to Mind Everywhere: An Experimentally-Grounded Framework for Understanding Diverse Bodies and Minds. *Front Syst Neurosci* **16**, 768201, doi:10.3389/fnsys.2022.768201 (2022).
91  Lyon, P. The biogenic approach to cognition. *Cogn Process* **7**, 11-29, doi:10.1007/s10339-005-0016-8 (2006).
92  Mathews, J. & Levin, M. The body electric 2.0: recent advances in developmental bioelectricity for regenerative and synthetic bioengineering. *Curr Opin Biotechnol* **52**, 134-144, doi:10.1016/j.copbio.2018.03.008 (2018).
93  LANGTON, C. *Proceedings of an interdisciplinary workshop on the synthesis and simulation of living systems*.  (ALIFE, 1989).